\pdfoutput=1

\documentclass[11pt]{article}

\usepackage[final]{acl}

\usepackage{times}
\usepackage{latexsym}

\usepackage[T1]{fontenc}

\usepackage[utf8]{inputenc}

\usepackage{microtype}

\usepackage{inconsolata}

\usepackage{graphicx}
\usepackage{amsmath,amssymb,amsfonts}%
\usepackage{amsthm}%
\usepackage{mathrsfs}%
\usepackage{xcolor}%
\usepackage{multirow}%
\usepackage{textcomp}%
\usepackage{manyfoot}%
\usepackage{booktabs}%
\usepackage{algorithm}%
\usepackage{algorithmicx}%
\usepackage{algpseudocode}%
\usepackage{listings}%
\usepackage{booktabs}
\usepackage{graphics}
\usepackage{colortbl}
\usepackage{MnSymbol}
\usepackage{wrapfig}
\usepackage{makecell}

\usepackage{fvextra}
\DefineVerbatimEnvironment{WrappedVerbatim}{Verbatim}{breaklines=true, fontsize=\small}

\definecolor{gold}{rgb}{1, 0.84, 0}
\definecolor{silver}{rgb}{0.75, 0.75, 0.75}
\definecolor{bronze}{rgb}{0.8, 0.5, 0.2}

\newcounter{example}
\definecolor{blackish1}{rgb}{0.3, 0.3, 0.3}
\definecolor{blackish2}{rgb}{0.1, 0.1, 0.1}

\newcommand\doc[2]{
   \leavevmode\par
   \stepcounter{example}
   \noindent
   \color{blackish1}
   \textbf{\theexample.) #1:} \\  #2\par }
   
\newcommand\pred[1]{
    \leavevmode\par\noindent                  
    \color{blackish2}
   {\leftskip10pt
     \textbf{Predicted keywords: }#1\par}}
     
\newcommand\true[1]{
    \leavevmode\par\noindent                  
    \color{blackish2}
   {\leftskip10pt
     \textbf{True keywords: }#1\par}}

\title{SEKE: Specialised Experts for Keyword Extraction}

\author{Matej Martinc \and Hanh Thi Hong Tran \and Senja Pollak \and Boshko Koloski \\
         Jožef Stefan Institute, Ljubljana, Slovenia \\
         \texttt{\{name\}.\{surname\}@ijs.si}}

\begin{document}
\maketitle
\begin{abstract}
Keyword extraction involves identifying the most descriptive words in a document, allowing automatic categorisation and summarisation of large quantities of diverse textual data. Relying on the insight that real-world keyword detection often requires handling of diverse content, we propose a novel supervised keyword extraction approach based on the  mixture of experts (MoE) technique. MoE uses a learnable routing sub-network to direct information to specialised experts, allowing them to specialise in distinct regions of the input space. SEKE, a mixture of Specialised Experts for supervised Keyword Extraction, uses DeBERTa as the backbone model and builds on the MoE framework, where experts attend to each token, by integrating it with a bidirectional Long short-term memory (BiLSTM) network, to allow successful extraction even on smaller corpora, where specialisation is harder due to lack of training data. The MoE framework also provides an insight into inner workings of individual experts, enhancing the explainability of the approach. We benchmark SEKE on multiple English datasets, achieving state-of-the-art performance compared to strong supervised and unsupervised baselines. Our analysis reveals that depending on data size and type, experts specialise in distinct syntactic and semantic components, such as punctuation, stopwords, parts-of-speech, or named entities. Code is available at \url{https://github.com/matejMartinc/SEKE_keyword_extraction}.
\end{abstract}

\section{Introduction}
\label{sec:intro}

    

Keyword extraction, i.e., extraction of words that represent crucial semantic aspects of the text, is crucial for efficiently summarizing, indexing, and retrieving relevant information from large textual corpora, making it one of the most fundamental NLP tasks \cite{song2023survey}. While the first automated solutions have been proposed more than two decades ago \citep{tumey1999learning,mihalcea2004textrank,hulth2003improved}, the task is far from solved, and the scientific community is still actively trying to improve the performance of the keyword extractors (see Section \ref{sec:related_works} for details).

Recently, the so-called foundation models have completely revolutionised the way natural language processing (NLP) is done, and the development of large language models (LLMs) has enabled task-agnostic architectures to solve downstream NLP tasks in a zero-shot fashion without the need for labelled data \citep{brown2020language}. While recent related studies \citep{martinez2023chatgpt,luo2023enhancing} and experiments conducted in this study show that LLMs' keyword extraction performance is still not comparable to specialised supervised approaches, their zero-shot performance nevertheless remains impressive. It was made possible, to a large extent, by the integration of novel training regimes and architectural components, such as Low-Rank Adaptation (LoRA) \cite{hu2021lora} and Mixture of Experts (MoE) \cite{jacobs1991adaptive}, which make models more efficient, while requiring fewer computational resources. While these components have been successfully employed in an unsupervised language modelling setting with abundant data, there are less studies exploring their effectiveness in a supervised setting with much fewer resources. The main reason for this is that these components were developed with a specific objective of improving the scalability of the models \cite{shazeer2017outrageously} and therefore their effect on the performance is somewhat neglected or measured only indirectly, that is, studies focus on how the use of these components allows for greater scalability, which in turn leads to performance gains.

In contrast, in this study, we focus on the performance improvements obtained from these components without increasing the size of the backbone model. The hypothesis we want to test is whether it is beneficial for keyword extraction to allow specific parts of the network to specialise for specific types of tokens. Since the ``keywordiness'' of a token is to a large extent determined by its specific contextual role, we explore if introducing a gating network that assigns different parts of the sequence to specialised layers, which would only attend to specific subsequences and tokens according to their semantic and grammatical role, could improve the current state-of-the-art (SOTA) in the field. 
In addition, we explore what type of specialisation different experts undertake and how much data is required for a successful specialisation. We test our approach on several datasets with different amounts of training data available, to determine the data threshold that still allows the convergence of expert layers. Finally, we examine whether synergy can be achieved between MoE and bidirectional Long short-term memory (BiLSTM) network that would lower this threshold and improve overall performance by amplifying the sequential information lost by the context slicing caused by the MoE strategy. The contributions of the paper are as follows.

\begin{figure}
    \centering
    \includegraphics[width=0.45\textwidth]{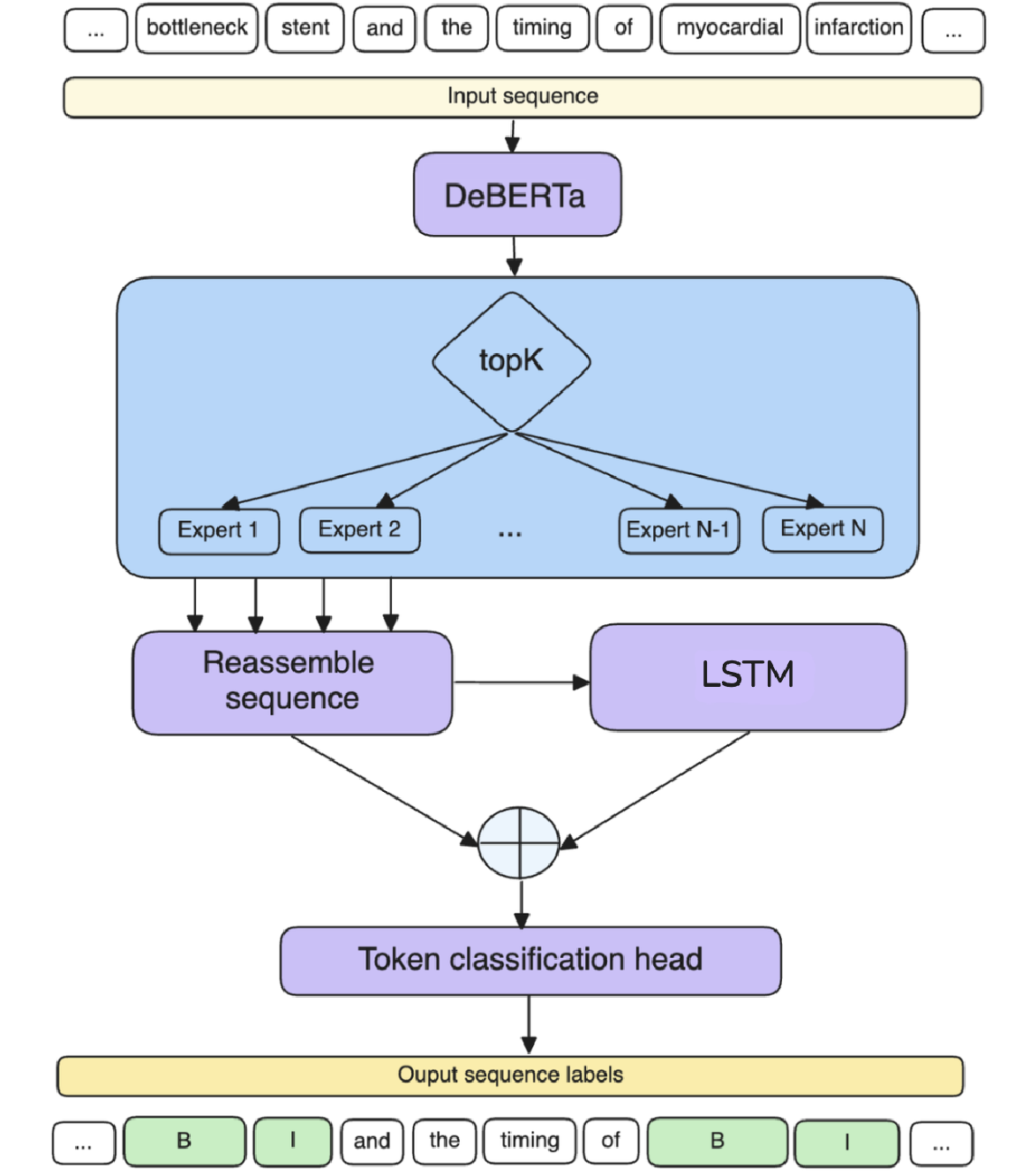}
    \caption{The general architecture given the input.}
    \label{fig:moe_architecture}
    \vspace{-1em}
\end{figure}

\begin{itemize}
    \item We propose a novel token classification head architecture (see Figure \ref{fig:moe_architecture}) that combines MoE and BiLSTM. The approach outperforms other SOTA keyword extraction approaches on most datasets. As far as we know, this is the first research that studies the usage of MoE in a supervised sequence-labelling scenario.
    \item We demonstrate that experts specialise in different syntactic and semantic structures, ranging from punctuation characters and part-of-speech tags to specific words and named entities, depending on the domain they are trained on and the size of the training data.
    \item  We show that adding MoE layers during fine-tuning generally has a positive effect on the model's performance no matter the dataset, but is somewhat dependent on the amount of data available. In addition, we show that the data requirements can be lowered by introducing additional BiLSTM layers.  
\end{itemize}

\vspace{-1em}

\section{Related Work}
\label{sec:related_works}

\subsection{Mixture of Experts}

The idea for Mixture of experts (MoE) comes from the study by \citet{jacobs1991adaptive}. The concept involved a supervised procedure for a system composed of separate networks, each responsible for a different subset of training cases. Each network, or expert, specialises in a distinct region of the input space. The choice of expert is managed by a gating network, which determines the weights for each expert. Both the experts and the gating network are trained during the training process.

More recently, the study by \citet{eigen2013learning} proposed integrating MoE as components within deep neural networks. This allowed MoE to function as layers in a multilayer network, enabling models to be both large and efficient, which led to the employment of the MoE method in the NLP setting. The study by \citet{shazeer2017outrageously} proposed a 137-billion-parameter LSTM model (the prevalent NLP architecture for language modelling at the time) by introducing sparsity, which allowed for very fast inference even at large scales. This work, which focused on translation, encountered several challenges such as high communication costs and training instabilities. After the success of the Mixtral 8x7B  \cite{jiang2024mixtral}, MoE have been used in several LLMs because they enable more efficient pre-training with fewer computing resources, allowing greater scalability in terms of model and data set size. The studies show that a MoE model achieves the same quality as its dense counterpart much faster during pre-training, leading to its usage in training multi-trillion parameter models \cite{fedus2022switch}. 

Although several studies showed that MoE are an efficient strategy to achieve greater scalability, very few studies (if any) focused on performance gains that could be obtained by this method. To our knowledge, no study to date explored how MoE can benefit supervised keyword extraction (or in fact any sequence labelling) tasks. 


\subsection{Keyword Extraction}
\label{sec:related_works_kw}

Contemporary studies on keyword extraction treat it either as a text generation or a sequence labelling task. \citet{yuan2019one} proposed an encoder-decoder RNN architecture featuring two mechanisms, \textit{semantic coverage} and \textit{orthogonal regularisation}, which treated keyword extraction as text generation. These mechanisms ensure that the representation of a generated keyword sequence is semantically aligned with the overall meaning of the source text. In contrast, several recent sequence labelling approaches tackle the task with transformers. \citet{sahrawat2019keyphrase} tested several transformer architectures, namely BERT \citep{devlin2018bert}, RoBERTa \citep{liu2019roberta}, and GPT-2, and two distinct sequence labelling heads, a BiLSTM and a BiLSTM head with an additional Conditional random fields layer (BiLSTM-CRF). The TNT-KID approach \cite{martinc2022tnt} on the other hand adapted the transformer architecture specifically for keyword extraction tasks by re-parameterizing the attention mechanism to focus more on positional information. Finally, \citet{sun2021capturing} proposed JointKPE, an open domain keyphrase extraction (KPE) architecture also based on pre-trained transformer language models. It ranks keyphrases by assessing their informativeness across the entire document and is simultaneously trained on a keyphrase chunking task to ensure the phraseness of the keyphrase candidates. 

Just recently, \citet{luo2023enhancing} proposed Diff-KPE, a text diffusion approach to generate enhanced keyphrase representations utilizing the supervised Variational Information Bottleneck (VIB). Diff-KPE generates keyphrase embeddings conditioned on the entire document using BERT and then injects the generated keyphrase embeddings into each phrase representation. 
Another recent approach is HybridMatch \cite{song2024match}, which consists of the representation-focused Siamese encoder component and the interaction-based component that estimates relatedness between candidate phrases and the corresponding document. 

Keyword extraction can also be tackled in an unsupervised manner. In general, these approaches can be divided into four main categories, namely statistical (e.g., YAKE \cite{campos2020yake}), graph-based (e.g., TextRank \cite{mihalcea2004textrank} and RaKUn \cite{vskrlj2019rakun}), embeddings-based (e.g., Key2Vec \cite{mahata2018key2vec}), and language model-based methods. Notably, large language model-based methods tend to be highly effective, driven by the rise of LLMs, such as ChatGPT \cite{brown2020language}. These models have been employed in several recent keyword extraction studies \cite{martinez2023chatgpt,luo2023enhancing,lee2023toward,peng2024model}, demonstrating promising results due to their zero-shot capabilities, which allow them to perform tasks without task-specific fine-tuning \cite{kojima2022large}. However, while significantly outperforming other unsupervised approaches, these models still do not offer performance comparable to the SOTA supervised methods. To some extent, this might be related to the deficiency observed by \citet{pmlr-v235-zhang24m}, which has highlighted the theoretical drawbacks of autoregressively trained LLM models for semantic classification tasks, attributing these limitations to their fixed token positioning, which restricts inter-sample connections. In contrast, masked language models, such as BERT \cite{devlin2018bert} and DeBERTa \cite{he2020deberta}, leverage flexible token targeting to overcome these issues. Additionally, through fine-tuning they can be easily (and cheaply) adapted to each specific keyword assignment regime, which vary across different domains. 

\section{Methods}
\label{sec:methods}

\subsection{Architecture}
\label{sec:architecture}

Our approach (visualised in Figure \ref{fig:moe_architecture}) is based on fine-tuning a pre-trained transformer architecture, in our case DeBERTa \cite{he2020deberta}, with a specialised token classification head. The architecture was chosen since it showcased a SOTA performance on several downstream NLP tasks, among them also sequence-labelling tasks such as NER \cite{shon2022slue}. Additionally, we hypothesised that the model is especially appropriate for keyword extraction due to its novel disentangled attention mechanism. In the DeBERTa model, each word is represented using two vectors that encode its content and position separately, and the attention weights among words are computed using disentangled matrices on their contents and relative positions, respectively. We assumed that this would allow the model to better distinguish between the positional and semantic/lexical information and therefore make it possible to assign attention to some tokens purely based on their position in the text, improving the overall performance. This hypothesis is based on a previous study by \citet{martinc2022tnt} showing that token position is especially important in the keyword identification task due to the disproportionate distribution of keywords in the document, which is skewed towards the beginning of the document (i.e., more keywords appear at the beginning of the document). By considering the importance of positional information, they fed the positional encoding to the attention mechanism directly, subsequently managing to improve the performance.

The main novelty of our approach is a customised token classification head with MoE. It is motivated by the hypothesis that token-based specialization offered by MoE might be especially useful for keyword extraction. The specialization of different attention heads and attention layers in the transformer architecture has been extensively analysed in the past, also in the context of keyword extraction \citep{martinc2022tnt}. This specialization is mostly embeddings-based, i.e., different parts of the embedding space for each token are specialized. In contrast, the specialization enabled by MoE occurs on the sequence level, since the router decides for each token in the sequence to which expert it should go. Since the ``keywordiness'' of a specific token (or a sequence of tokens) to a large extent depends on a context, we assumed that this type of specialization might bring benefits. 

The output logits from the DeBERTa backbone are first fed to the gate network or router that determines which tokens in the input sequence are sent to which expert. More specifically, in the setting with \textit{n} experts, the gate network (G) decides which experts (E) receive a part of the input:

\begin{equation*}
    y = \sum_{i=1}^{n}G(x)_iE_i(x)
\end{equation*}

We model the routing as a weighted multiplication, where hypothetically all experts could contribute to processing each part of the input. Nevertheless, the additional \textit{top\_k} parameter ensures that only top  \textit{k} experts contribute to the respective part of the input, and other experts are excluded from the computational graph, hence saving computational resources. Following \citet{shazeer2017outrageously}, we employ the so-called Noisy Top-k Gating, which introduces some (tunable) noise and then keeps the top k values. Noise is introduced for load balancing, i.e., to prevent the gating network from converging in a way that the same few experts are activated for most of the tokens, which would make training inefficient. The final weight distribution is obtained by feeding the input ($x$) through a dense layer ($W_g$) and adding some random noise in the following way: 

\begin{equation*}
    H(x)_i = (x \cdot W_g)_i + \textrm{RND} \cdot \textrm{Softplus}((x \cdot W_{noise})_i)
\end{equation*}

Here, RND represents a weight vector of random numbers from a normal distribution with mean 0 and variance 1.
After that, we filter out only the top k experts out of v:

\begin{equation*}
KeepTopK(v,k)_i=\{v_i \textrm{ if } v_i  \textrm{ in top k, else} -\inf\}
\end{equation*}

We finally apply the softmax to filtered experts:

\begin{equation*}
G(x)=\textrm{Softmax}(\textrm{KeepTopK}(H(x),k))
\end{equation*}

Note that the router is composed of learnable parameters and is fine-tuned at the same time as the rest of the network. On the other hand, each expert is represented as a neural network consisting of three sequential dense layers $W_g$ with additional activation and dropout layers. More specifically, the expert is represented as:

\begin{equation*}
E(x)= \textrm{Dropout}(\textrm{ReLU}(x \cdot W_g) \cdot W_g) \cdot W_g
\end{equation*}

The outputs of different experts, which attend to specific tokens assigned to them by the routing network, are first reassembled into a single sequence representation. Then, a two-layer randomly initialised encoder, consisting of dropout and two BiLSTM layers, is added (with element-wise summation) to this sequence. The employment of the BiLSTM module is there to amplify the sequential information lost by the context slicing caused by the MoE strategy, i.e., the weakening of the positional information due to overspecialization. This is especially problematic in case of multi-word keyphrases, where different parts of the keyphrase are attended by different experts. This adaptation is also motivated by the findings from the related work \cite{sahrawat2019keyphrase}, where it was shown that putting a BiLSTM classification head on top of different transformer models improved the keyword extraction capabilities of the model.


The output sequence of the MoE and BiLSTM encoders is finally fed to the feed-forward classification layer, which returns the output matrix of size $\textrm{SL} * \textrm{NC}$, where SL stands for sequence length and NC stands for the number of classes. In our case, the $NC$ is 3, since we model the keyword extraction task as sequence labelling and employ the so-called BIO labelling approach (same as, for example, in \citet{sahrawat2019keyphrase}). Using the BIO annotation regime, each word in the document is assigned one of the three possible labels: $k_b$ denotes that the word is the first word in a keyphrase, $k_i$ means that the word is inside a keyphrase, and $k_o$ indicates that the word is not part of a keyphrase. 

The identified keyphrases are extracted from the labelled sequence produced by the model, and we conduct additional filtering, where we remove keyphrases containing punctuation (except dashes and apostrophes), and deduplication, where we remove duplicate keyphrases. The final output list contains a maximum of 10 keyphrases per input document. 

\section{Experimental Setting}

\subsection{Datasets}
\label{sec:datasets}
We conduct experiments on selected benchmark datasets used in the SOTA papers to which we compare our approach. More specifically, we experiment on 6 publicly available datasets with MIT license from the related work \cite{meng2017deep,martinc2022tnt,xiong2019open}, covering three domains, news, scientific articles, and web pages.

We present the datasets in Table \ref{tab:data_all}, which contains information about the domain, the number of train, validation, and test documents, as well as average number of keywords per document. There is notable variability in a number of keywords per document, with some datasets requiring extraction of twice (or even three times) as many keywords as others.

\begin{table}[h!]
    \centering
    \resizebox{0.47\textwidth}{!}{
    \begin{tabular}{l|c|r|r|r|r|r|c}
        \toprule
        \midrule
        \bf Dataset & \bf Domain & \bf Train &\bf Valid &\bf Test & \bf Train & \bf Valid \ & \bf Test \ \\                 & & & \bf  \#Docs &  \bf \#Docs & \bf \#KW & \bf \#KW & \bf \#KW \\
        & & & &  & \bf per doc & \bf per doc & \bf per doc \\
        \midrule
        \bf KP20k & \bf CS & 530,000 & 20,000 & 20,000 &  3.30 & 3.31 & 3.32 \\ 
        \bf Inspec &\bf CS & / & 1,500 & 500 & / & 7.21 & 7.51  \\
        \bf Krapivin &\bf CS & / & 1,844 & 460 & / & 2.87 & 3.22 \\ 
        \bf KPTimes &\bf News & 259,923 & 10,000  & 10,000  & 2.34 & 2.31 & 2.35 \\ 
        \bf JPTimes &\bf News & - & - & 10,000 & / & / &  3.84  \\
        \bf openKP &\bf Web & 134,894 & 6,616 & 6,614 & 2.00 & 1.90  & 1.92 \\
        \midrule
        \bottomrule
    \end{tabular}
    }
    \caption{Statistics for keyword extraction datasets. This table presents the number of documents in the train, validation, and test sets for each dataset. The \textbf{Domain} column indicates the source of the datasets (computer science (CS), news, or web), while \textbf{\#KW per doc} shows the average number of keywords per document.
    }
    \label{tab:data_all}
    \vspace{-0.3cm}
\end{table}

\subsection{Experiments}

We test different settings to determine the effect that the proposed architectural additions have on the performance of the model. We are interested in the specific contributions of the MoE and BiLSTM layers to the performance of the model (either separately or together); therefore, we compare the settings containing these components to the baseline DeBERTa model with just a (usual) feed-forward token classification head.  

We employ the DeBERTa model\footnote{More specifically, the ``deberta-v3-base'' version available at \url{https://huggingface.co/microsoft/deberta-v3-base} under the MIT license.} and fine-tune it using Low-Rank Adaptation (LoRA) \cite{hu2021lora}. We freeze all layers in the backbone model and train only low-rank perturbations to query and value weight matrices in the model. Additionally, we train all the matrices in the custom token classification head (i.e., MoE, BiLSTM, and dense classification layers are randomly initialised and fine-tuned). See Section \ref{sec:appendix-hyperparameters} for details about the hyperparameter setting used.

In order to determine how the size of the training dataset affects our approach, we employ two distinct training scenarios. In the \textbf{validation set training} scenario, the DeBERTa model was fine-tuned just on each dataset's validation sets, which were randomly split into 80 percent of documents used for fine-tuning and 20 percent of documents used for hyperparameter optimisation and test set model selection. We fine-tune the models for up to 20 epochs and employ early stopping. Note that the \textit{JPTimes} dataset with no available predefined validation-test splits is only used for testing. Here, the model was fine-tuned on the \textit{KPTimes-valid} dataset.

In the \textbf{full training} scenario, we fine-tune the models on all available data, same as in related work \cite{meng2017deep, martinc2022tnt}. More specifically, before fine-tuning the model on three relatively small validation datasets containing scientific papers (\textit{KP20k-valid}, \textit{Inspec-valid}, and \textit{Krapivin-valid}), we first ``pre-train'' the model (using the same token classification objective as in the validation dataset fine-tuning stage) for 20 epochs on the large \textit{KP20k-train} dataset also containing scientific articles. We do the same for news datasets, i.e., before fine-tuning the model on the \textit{KPTimes-valid} dataset, we ``pre-train'' it on the large \textit{KPTimes-train} dataset. For the \textit{JPTimes} dataset with no available predefined validation-test splits, the keyword detection model was ``pre-trained'' on \textit{KPTimes-train} and fine-tuned on the \textit{KPTimes-valid} dataset. For the openKP dataset, the model is ``pre-trained'' on the \textit{openKP-train} dataset.

To assess the performance of our models and compare the results of our keyword extraction approach to other SOTA methods, we adopt the same evaluation methodology as in \citet{meng2017deep,martinc2022tnt,luo2023enhancing}. We use the same test splits for the sake of comparability between our approach and others, and measure F1@$k$ with $k \in \{1,3,5\}$ on the \textit{openKP} dataset and $k \in \{5,10\}$ on other datasets\footnote{Note that the values of $k$ differ between \textit{openKP} and other datasets. This is due to the variability of keyword assignment regimes across different domains, i.e., news and scientific articles on average have more keywords than web pages (see Table \ref{tab:data_all} for details). We report the same F1@$k$ for each dataset as in related work to facilitate easy comparison.}. To determine the exact match, we first lowercase the candidate keywords and the gold standard keywords, and then apply Porter Stemmer \cite{porter1980algorithm} to both.

\begin{table*}[ht!]
    \centering
    \resizebox{1.0\textwidth}{!}{
    \begin{tabular}{c|cc|cc|cc|cc|cc|ccc}
    \toprule
    \midrule
    \multirow{3}{*}{}& \multicolumn{2}{c|}{\underline{KP20k}} & \multicolumn{2}{c|}{\underline{Inspec}} & \multicolumn{2}{c|}{\underline{Krapivin}} & \multicolumn{2}{c|}{\underline{KPTimes}} & \multicolumn{2}{c}{\underline{JPTimes}} & \multicolumn{3}{c}{\underline{openKP}} \\ 
    & \bf F1@5 & \bf F1@10 & \bf F1@5 & \bf F1@10 & \bf F1@5 & \bf F1@10 & \bf F1@5 & \bf F1@10 & \bf F1@5 & \bf F1@10 & \bf F1@1 & \bf F1@3 & \bf F1@5 \\
    \midrule \midrule
    \multicolumn{14}{c}{\textbf{ Unsupervised baselines }} \\\midrule \midrule
    TF-IDF & 7.2 & 9.4 & 16.0 & 24.4 & 6.7 & 9.3 & 17.9 & 15.1 & 26.6 & 22.9 & 19.6 & 22.3 & 19.6 \\
    TextRank & 18.0 & 15.0 & 28.6 & 33.9 & 18.5 & 16.0 & 2.2 & 3.0 & 1.2 & 2.6 & 5.4 & 7.6 & 7.9 \\
    RaKUN & 17.7 & 16.0 & 10.1 & 10.8 & 12.7 & 10.6 & 16.8 & 13.9 & 22.5 & 18.5 & 8.81 & 10.12 & 6.45 \\
    YAKE & 14.1 & 14.6 & 20.4 & 22.3 & 21.5 & 19.6 & 10.5 & 11.8 & 10.9 & 13.5 & 3.73 & 4.24 & 2.94  \\ 
    \midrule \midrule \multicolumn{14}{c}{\textbf{ Large Language Model baselines }} \\\midrule \midrule
    ChatGLM2-6b & / & / & 25.1 & 30.1 & / & / & / & / & / & / & 16.0 & 11.0 & 8.6 \\

    ChatGPT & 23.2 & 10.8 & 35.2 & 33.9 & 12.8 & 11.7 & 27.9 & / & / & / & 20.8 & 20.4 & 16.6 \\ 
    GPT4o-mini & 22.7 & 19.8 & 34.4 & 42.1 & 22.1 & 20.5 & 19.5 & 16.6 & 29.3 & 25.2  & 34.0 & 26.8 & 26.8 \\ 
    Gemini-1.5-flash-8B & 26.5 & 23.6 & 35.9 & 42.1 & 26.6 & 25.9 & 20.4 & 26.9  & 31.8 & 27.5 & 33.1 & 25.5 & 25.6 \\
    \midrule \midrule \multicolumn{14}{c}{\textbf{ Supervised baselines }} \\\midrule \midrule
    CopyRNN & 31.7 & 27.3 & 24.4 & 28.9 & 30.5 & 26.6 & 40.6 & 39.3 & 25.6 & 24.6 & 21.7 & 23.7 & 21.0 \\
    CatSeqD & 34.8 & 29.8 & 27.6 & 33.3 & 32.5 & 28.5 & 42.4 & 42.4 & 23.8 & 23.8 & / & / & / \\
    GPT-2 & 27.5 & 27.8 & 41.3 & 46.9 & 25.3 & 25.3 & 42.1 & 42.3 & 33.1 & 33.6 & / & / & / \\
    GPT-2+BiLSTM-CRF & 35.5 & 36.0 & 46.2 & 52.4 & 28.7 & 28.8 & 47.8 & 47.9 & 38.6 & 38.9 & / & / & / \\
    TNT-KID & 33.6 & 33.8 & 46.0 & 53.6 & 31.0 & 32.0 & 48.5 & 48.5 & 35.9 & 36.1 & / & / & / \\
    JointKPE & \textbf{41.7} & 34.4 & 35.2 & 35.0 & \textbf{36.0} & 29.2 & / & / & / & / & 36.4 & 39.1 & 33.8 \\
    Diff-KPE & \textbf{41.7} & 34.3 & 32.3 & 35.0 & 35.0 & 31.4 & / & / & / & / & \textbf{37.8} & 38.5 & 32.7 \\
    HybridMatch & / & / & / & / & / & / & / & / & / & / & 37.3 & 39.4 & 34.1 \\
    \midrule \midrule
    \multicolumn{14}{c}{\textbf{ Our approaches - validation set training }} \\ \midrule \midrule
    DeBERTa & 29.2±1.3 & 29.4±1.3 & 44.8±0.6 & 53.9±1.1 & 26.8±1.1 & 27.5±1.0 & 41.9±0.7 & 41.9±0.7 & 35.2±0.9 & 35.4±0.9 & 27.6±1.5 & 36.5±0.9 & 37.2±0.8 \\
    DeBERTa MoE & 31.1±1.0 & 31.5±1.1 & 44.1±0.7 & 53.7±0.5 & 26.9±0.3 & 27.4±0.5 & 42.5±0.4 & 42.5±0.4 & 36.4±2.4 & 36.8±2.6 & 27.7±1.0 & 38.0±0.7 & 38.9±0.6 \\
    DeBERTa BiLSTM & 31.2±1.0 & 31.5±1.0 & 47.7±0.7 & 56.6±0.8 & 29.2±1.5 & 29.6±1.5 & 44.1±0.7 & 44.2±0.7 & 36.2±1.2 & 36.6±1.3 & 26.8±1.1 & 37.2±0.5 & 38.1±0.4 \\
    DeBERTa MoE BiLSTM & 32.7±1.1 & 33.0±1.1 & 47.2±1.1 & 56.5±1.0 & 29.5±1.1 & 30.1±1.0 & 44.9±0.3 & 45.1±0.3 & 37.4±0.5 & 37.8±0.6 & 27.1±0.9 & 38.9±0.5 & 40.1±0.4 \\  \midrule \midrule
    \multicolumn{14}{c}{\textbf{ Our approaches - full training }} \\ \midrule \midrule
    DeBERTa & 34.7±0.6 & 35.1±0.6 & 48.1±0.8 & 57.4±0.8 & 31.8±1.9 & 32.1±1.9 & 54.5±0.6 & 54.5±0.6 & 37.5±1.4 & 37.7±1.4 & 33.7±0.9 & 41.4±1.2 & 41.7±1.2 \\
    DeBERTa MoE & 33.5±0.7 & 33.9±0.6 & 47.1±0.5 & 56.3±1.1 & 30.0±0.8 & 30.2±0.8 & \textbf{55.0±0.4} & \textbf{55.1±0.4} & 38.7±0.9 & 39.1±1.0 & 34.2±0.4 & 42.4±0.4 & 42.8±0.4 \\
    DeBERTa BiLSTM & 34.4±0.6 & 34.7±0.6 & 47.6±0.6 & 57.5±0.8 & 32.3±1.5 & 32.6±1.4 & 53.5±0.1 & 53.5±0.1 & 38.1±0.9 & 38.2±1.0 & 31.9±2.3 & 39.8±1.0 & 40.2±0.8 \\
    DeBERTa MoE BiLSTM & 36.0±0.3 & \textbf{36.5±0.3} & \textbf{49.4±1.0} & \textbf{58.8±0.4} & 34.3±0.9 & \textbf{34.7±0.9} & 54.8±1.0 & 55.0±0.9 & \textbf{39.0±1.7} & \textbf{39.3±1.8} & 33.3±1.5 & \textbf{42.5±0.6} & \textbf{42.9±0.5} \\

    \midrule
    \bottomrule
    \end{tabular}
    }
     \caption{Empirical evaluation of the keyword extractors. We report the average over five runs and the standard deviation. Bold represents the best score according to a specific criterion on a specific dataset.}
    \label{tbl:results}
    \vspace{-1em}
\end{table*}

\section{Results and Analysis}
\label{sec:results}

In Table \ref{tbl:results}, we present the results of our approach and compare them to several baselines (see Section \ref{sec:related_works_kw} for details). TF-IDF, TextRank, RaKUN and YAKE algorithms are unsupervised and do not require any training. Same applies to the LLM-based approaches, namely ChatGPT\footnote{We report results for the gpt-3.5-turbo version tested in \citet{luo2023enhancing}.}, ChatGLM2-6B\footnote{We list results from \citet{luo2023enhancing}.}, GPT4o-mini \citep{achiam2023gpt} and Gemini-1.5-flash-8B \citep{team2023gemini}\footnote{see Section \ref{sec:prompts} for details about the LLM experiments.}.

For the supervised baselines, CopyRNN, CatSeqD, GPT-2, GPT-2+BiLSTM-CRF, and TNT-KID, we list results reported by \citet{martinc2022tnt}, for Diff-KPE, we list results reported by \citet{luo2023enhancing}, for JointKPE, we list results from \citet{sun2021capturing}, and for HybridMatch, we list results from \citet{song2024match}. Note that GPT-2, GPT-2 + BiLSTM-CRF and TNT-KID are sequence labelling approaches, same as ours. Since they return a non-predetermined number of keywords (i.e., the approaches learn during training how many keywords they should return), they require adaptation to each specific keyword labelling regime for optimal performance and were therefore fine-tuned on the validation datasets. In contrast, JointKPE, DIffKPE nad HybridMatch baselines return a predetermined number of best ranked candidates (3, 5 or 10, depending on the evaluation criteria) and do not require validation set fine-tuning for adaptation. Same applies to the two generation baselines, CopyRNN and CatSeqD, which tend to generate a large set of predicted keywords, from which  a fixed number of top predicted phrases (3, 5 or 10) is taken to compare with the ground truth. Finally, besides testing our custom sequence labelling heads (containing MoE, BiLSTM, or both), we also report results for the unmodified pre-trained DeBERTa model with a standard feed-forward token classification head. 

In the \textbf{validation set training scenario}, by adding the MoE on top of the DeBERTa model (see configuration DeBERTa MoE), we improve the performance of the vanilla model (see configuration DeBERTa) with a feed-forward token classification head on four out of six datasets according to all evaluation criteria. The improvement is not observed for the two smallest datasets, \textit{Krapivin} and \textit{Inspec}. On the other hand, adding the additional BiLSTM head to the vanilla DeBERTa model improves the performance of the base model on all datasets (see configuration DeBERTa BiLSTM) and according to almost all criteria (the only exception being the F1@1 criterion on the \textit{openKP} dataset). The results therefore indicate a synergy between the MoE and BiLSTM layers, since the model with both MoE and BiLSTM layers manages to substantially improve the base model and also outperforms the DeBERTa MoE and DeBERTa BiLSTM models on four out of six datasets according to all criteria. 

In the \textbf{full training scenario}, the DeBERTa MoE model outperforms the vanilla DeBERTa on three out of six datasets according to all evaluation criteria. Interestingly, the DeBERTa MoE model fails on all computer science datasets, which suggests that expert specialisation might be harder on this domain. On the other hand, the DeBERTa MoE BiLSTM model outperforms the vanilla one on all but one dataset and criterion, the F1@1 criterion on the \textit{openKP} dataset. 

In this scenario, we also manage to substantially improve the SOTA on the \textit{Inspec}, \textit{KPTimes}, and \textit{JPTimes} datasets according to both criteria. While we outperform the best-performing model \textit{JoinKPE} in F1@10 on the \textit{KP20k} and the \textit{Krapivin} datasets, our best-performing configuration (DeBERTa MoE BiLSTM) on the other hand achieves uncompetitive performance in terms of F1@5 on these two datasets. On the \textit{openKP}, our best approach (DeBERTa MoE BiLSTM) fails to beat several approaches (among others, also the two LLM-based baselines, GPT4o-mini and Gemini-1.5-flash-8B) according to the F1@1 criterion. The performance comparison is nevertheless favourable according to the other two criteria. 

The results also show that the unsupervised models in general achieve much worse performance than the supervised models. While the LLM-based approaches outperform other unsupervised baselines, their performance lags behind supervised approaches. Out of the four tested LLMs, the two proprietary models, GPT4o-mini and Gemini-1.5-flash-8B, perform the best, with Gemini-1.5-flash-8B being the best model on most datasets. 

\subsection{Examining the Specialisation of Experts}

Next, we assess how individual experts correlate with the tokens they operate on to determine their specific types of \textbf{specialisation}. During inference on the test sets, we extracted the expert with the highest weight (i.e., the expert contributing the most out of the top $k$) for each token in the input sequence. We then correlated the selected expert per token with the token's grammatical and semantic attributes. On the grammatical/lexical level, we consider: whether a token is a specific word, part-of-speech tag (\textbf{POS}), punctuation character or not (\textbf{Punct}), or stopword or not (\textbf{Stop.}). On the semantic level, we examine the predicted labels for each token (\textbf{Labels}), whether a given token is a named entity (\textbf{bin. NE}), and if so, the specific type of named entity (\textbf{NE}). This way, we try to identify different types of \textbf{specialisation}, i.e., whether specialisation is rooted in syntax or semantics.

We annotated the tokens with POS tags and NE labels using the SpaCy library~\cite{honnibal2020spacy}. In total, we obtain six different categories of specialisation, which we annotate on the dataset level (i.e., we obtain annotations for the concatenation of documents belonging to a specific dataset). Since we work with categorical values, we measure the correlation between the highest-weighted experts and each category by employing \textit{Cramér's V} statistic~\cite{cramer1999mathematical} with the bias correction proposed in \citet{bergsma2013bias}. We consider the two training regimes, validation set and full training, separately, to examine the impact of data size. We present the results in Table~\ref{tab:ablation-kw}.

\begin{table}[h!]
    \centering
    \resizebox{0.48\textwidth}{!}{
    \begin{tabular}{l|r|r|r|r|r|r}
        \toprule
        \midrule
       \textbf{Experts to} & \textbf{Inspec} & \textbf{Krapivin} & \textbf{KP20k} & \textbf{KPTimes} & \textbf{JPTimes} & \textbf{openKP} \\ \midrule
        \multicolumn{7}{c}{\textbf{ Validation set training }} \\\midrule
        \textbf{Words} & \cellcolor{bronze} 0.609 & \cellcolor{gold}0.456 &  \cellcolor{silver}  0.592 & \cellcolor{gold}0.465 & \cellcolor{gold} 0.525 & \cellcolor{gold} 0.475 \\
        \textbf{Punct.} & \cellcolor{gold} 0.701 &  \cellcolor{bronze} 0.391 & \cellcolor{gold} 0.688 & 0.076 & 0.063 & 0.000 \\
        \textbf{Stop.} & \cellcolor{silver} 0.614 & \cellcolor{silver} 0.403 & \cellcolor{bronze} 0.555 & 0.160 & 0.096 & \cellcolor{silver}  0.339 \\
        \textbf{POS}  & 0.571 & 0.376 & 0.528 &  \cellcolor{bronze} 0.260 & \cellcolor{silver}   0.323 & \cellcolor{bronze} 0.305 \\
        \textbf{Labels} & 0.257 & 0.184 & 0.135 & 0.093 & 0.107 & 0.153 \\
        \textbf{bin. NE} & 0.085 & 0.025 & 0.056 & \cellcolor{silver} 0.271 & 0.198 & 0.148 \\
        \textbf{NE} & 0.070 & 0.045 & 0.046 & 0.246 & \cellcolor{bronze} 0.286 & 0.162 \\\midrule
        \multicolumn{7}{c}{\textbf{ Full training }} \\\midrule
        \textbf{Words} & \cellcolor{gold} 0.575 & \cellcolor{gold} 0.468 & \cellcolor{gold}  0.459 & \cellcolor{gold} 0.465 & \cellcolor{gold} 0.367 & \cellcolor{gold} 0.521 \\
        \textbf{Punct.} & 0.128 & 0.081 & 0.048 & 0.082 & 0.040 & 0.009 \\
        \textbf{Stop.} & \cellcolor{bronze}  0.257 & 0.105 & 0.115 & 0.168 & 0.092 & 0.224 \\
        \textbf{POS} & \cellcolor{silver} 0.415 & \cellcolor{silver} 0.265 & \cellcolor{bronze}  0.262 & \cellcolor{silver} 0.267 & \cellcolor{silver} 0.165 & \cellcolor{bronze} 0.255 \\
        \textbf{Labels}  & 0.228 & \cellcolor{bronze}  0.120 & \cellcolor{silver} 0.341 & 0.184 & 0.089 & \cellcolor{silver} 0.269 \\
        \textbf{bin. NE} & 0.039 & 0.029 & 0.043 & \cellcolor{bronze} 0.238 & 0.131 &   0.171 \\
        \textbf{NE} & 0.048 & 0.028 & 0.052 & 0.237 & \cellcolor{bronze} 0.139 & 0.167 \\\midrule
        \bottomrule
    \end{tabular}
    }
    \caption{Results of the Cramer's V statistic for correlation between the most weighted expert for a specific token and the specialisation categories examined, across \textbf{Validation set training} and \textbf{Full training} regimes. \textcolor{gold}{Strongest}, \textcolor{silver}{second strongest}, and \textcolor{bronze}{third strongest} correlations per dataset and regime are marked.} 
    \label{tab:ablation-kw}
    \vspace{-0.4cm}
\end{table}

In the validation set training scenario, a strong correlation between specific words and specific experts is observed on all datasets. This is not surprising, since MoE works on the token level. In contrast, the correlation between experts and labels (I, O, and B) tends to be moderate at best (i.e., between 0.1 and 0.3) on most datasets. One can also observe a moderate (and on some datasets strong, i.e., more than 0.5) correlation between specific experts and POS tags, suggesting that specialisation is somewhat influenced by the syntax of the text. 

By looking at the other three types of correlation, namely the correlation between experts and NEs, between experts and stopword tokens, and between experts and punctuation tokens, one can see a very distinct difference between the computer science and other datasets. While there is a strong tendency for experts to specialise for stopword and punctuation tokens on all the computer science datasets (\textit{Inspec}, \textit{Krapivin}, and \textit{KP20k}), this tendency is much weaker on other (news and web page) datasets, where one can observe almost no correlation. Here one can observe a specialisation of experts for NEs, which is almost non-existent on the computer science datasets. A correlation is almost the same if we binarize the NE sequence (i.e., to NE/non-NE tokens) or if we look for correlations between experts and 23 labels returned by the SpaCy NE recognition pipeline. The specialisation for NEs can be perhaps explained by the fact that news and web datasets contain much more NEs than scientific papers and that many of them are in fact labelled as keywords. The strong correlation with simple syntactic patterns on the computer science datasets can be on the other hand explained by the lack of obvious semantic clues, such as NEs, in these datasets.

When comparing the specialisation in the validation set training scenario and the full training scenario, one can see that correlation to labels, words, NEs, and binary NEs is mostly the same. On the other hand, one can see a decrease in correlation to POS tags, stopwords and punctuation on the computer science datasets, when the models are trained on more data. The assumption is that this is the consequence of expert specialisation becoming more complex when the amount of data allows it. 

\subsection{Error Analysis of MoE Performance by Keyword Length}

\begin{table*}[h!]
\centering
\resizebox{1.0\textwidth}{!}{
\begin{tabular}{lcccccccccc}
\toprule
\textbf{Dataset/Keyword Length} & \textbf{1} & \textbf{2} & \textbf{3} & \textbf{4} & \textbf{5} & \textbf{6} & \textbf{7} & \textbf{8} & \textbf{9} & \textbf{10} \\
\midrule
\textbf{Inspec} & 
\textbf{\makecell{0.0018 \\ (n=291)}} & \makecell{-0.0042 \\ (n=472)} & \textbf{\makecell{0.0136 \\ (n=396)}} & \textbf{\makecell{0.0303 \\ (n=209)}} & \textbf{\makecell{0.0353 \\ (n=85)}} & \textbf{\makecell{0.2610 \\ (n=23)}} & \textbf{\makecell{0.1004 \\ (n=10)}} & \textbf{\makecell{0.2381 \\ (n=3)}} & \makecell{0.0000 \\ (n=1)} & \makecell{0.0000 \\ (n=1)} \\
\addlinespace 
\textbf{JPTimes} & 
\textbf{\makecell{0.0175 \\ (n=9241)}} & \makecell{-0.0386 \\ (n=6642)} & \makecell{-0.0362 \\ (n=1576)} & \textbf{\makecell{0.0036 \\ (n=274)}} & \textbf{\makecell{0.0417 \\ (n=48)}} & \textbf{\makecell{0.1333 \\ (n=15)}} & \textbf{\makecell{0.2500 \\ (n=4)}} & \makecell{0.0000 \\ (n=4)} & --- & \makecell{0.0000 \\ (n=1)} \\
\addlinespace
\textbf{KP20k} & 
\makecell{-0.0196 \\ (n=9479)} & \textbf{\makecell{0.0035 \\ (n=14905)}} & \textbf{\makecell{0.0202 \\ (n=7943)}} & \textbf{\makecell{0.0675 \\ (n=2040)}} & \textbf{\makecell{0.0527 \\ (n=450)}} & \textbf{\makecell{0.0103 \\ (n=97)}} & \makecell{0.0000 \\ (n=27)} & \makecell{0.0000 \\ (n=6)} & \makecell{0.0000 \\ (n=5)} & --- \\
\addlinespace
\textbf{KPTimes} & 
\textbf{\makecell{0.0222 \\ (n=7184)}} & \textbf{\makecell{0.0814 \\ (n=5027)}} & \textbf{\makecell{0.1121 \\ (n=1766)}} & \textbf{\makecell{0.1569 \\ (n=465)}} & \textbf{\makecell{0.1192 \\ (n=151)}} & \textbf{\makecell{0.0909 \\ (n=88)}} & \textbf{\makecell{0.3824 \\ (n=34)}} & \textbf{\makecell{0.5556 \\ (n=9)}} & \makecell{0.0000 \\ (n=1)} & \makecell{0.0000 \\ (n=2)} \\
\addlinespace
\textbf{Krapivin} & 
\textbf{\makecell{0.0002 \\ (n=206)}} & \textbf{\makecell{0.0139 \\ (n=381)}} & \textbf{\makecell{0.0273 \\ (n=174)}} & \textbf{\makecell{0.0063 \\ (n=50)}} & \makecell{0.0000 \\ (n=8)} & \makecell{0.0000 \\ (n=6)} & --- & --- & --- & --- \\
\addlinespace
\textbf{openKP} & 
\textbf{\makecell{0.0180 \\ (n=3125)}} & \textbf{\makecell{0.0311 \\ (n=4118)}} & \textbf{\makecell{0.0211 \\ (n=1956)}} & \textbf{\makecell{0.0159 \\ (n=599)}} & \textbf{\makecell{0.0072 \\ (n=139)}} & \textbf{\makecell{0.0385 \\ (n=26)}} & \makecell{0.0000 \\ (n=5)} & \makecell{0.0000 \\ (n=4)} & \makecell{0.0000 \\ (n=1)} & --- \\
\bottomrule
\end{tabular}
}
\caption{Difference in F1@1 scores between DeBERTa MoE BiLSTM and DeBERTa MoE (DeBERTa MoE BiLSTM F1@1 - DeBERTa MoE F1@1) on keywords of different word length across different datasets.}
\label{tab:keyword_analysis_multiline}
\end{table*}

To test the hypothesis that the MoE component struggles with longer keywords due to context slicing, we have conducted an additional error analysis. This ablation study compares the performance of DeBERTa MoE BiLSTM and DeBERTa MoE model on keywords of varying lengths.

Specifically, for each test dataset, we created subsets of documents containing keywords of a specific length, ranging from one to ten words. We then evaluated the performance in terms of F1@1 score for each model on these subsets, considering only the keywords of that specific length (we use F1@1 due to small set of keywords of specific length per document).

Table \ref{tab:keyword_analysis_multiline} presents the differences in F1@1 scores (calculated as DeBERTa MoE BiLSTM F1@1 - DeBERTa MoE F1@1). Positive values indicate that DeBERTa MoE BiLSTM outperformed DeBERTa MoE on keywords of specific length, while negative values indicate the opposite. The results show that the standard MoE model is indeed less effective at extracting long, multi-word keywords compared to the model with an BiLSTM encoder (i.e., the difference is generally bigger on longer keywords). This analysis supports the hypothesis that context slicing in the MoE architecture is a key deficiency.

\section{Conclusion}
\label{sec:conclusion}
In summary, we investigated the potential of MoE for keyword extraction tasks. We introduced SEKE, a novel architecture that combines MoE and BiLSTM built upon the DeBERTa model. This approach achieved superior performance compared to existing SOTA models on standard datasets for keyword extraction. The study revealed that adding BiLSTM can help mitigate the data limitations, showing the feasibility of the approach even on small training datasets. In the future, we will try to pinpoint the exact dataset size thresholds, down to which the proposed approach is still feasible. Additionally, we will expand the evaluation to more benchmarks to further strengthen the analysis.

In this work, our primary goal was to evaluate the general effectiveness of our proposed framework using a consistent and reasonable configuration across all experiments. To maintain a fair comparison and avoid overfitting to specific datasets, we used commonly adopted default values (e.g., 4 experts, top-k = 2) without extensive hyperparameter tuning. In the future, a more exhaustive hyperparameter search will be conducted, which could yield additional performance gains and further insights into the model behaviour. 

The analysis revealed that specialisation is domain-specific. Current results indicate that in text with higher complexity (i.e., scientific papers) experts to a larger extent rely on syntactical patterns than in the less complex text, where experts to a larger extent specialise for processing of semantic clues. This hypothesis will be further tested in the future by applying the approach in new domains and languages other than English. 

Finally, we can observe that despite the impressive improvements in the development of foundational LLMs, keyword extraction task can still not be sufficiently solved by employing these models, which lag behind supervised approaches on most datasets. Since the biggest advantage of supervised approaches is their ability to adapt to the specifics of the syntax, semantics, content, and keyword tagging regime of the specific corpus \cite{meng2017deep}, in the future we will adapt LLMs to the specific corpora and keyword extraction regimes with fine-tuning and in-context learning.

\section{Limitations}

First, the empirical evaluation was conducted exclusively on English-language datasets from three specific domains (computer science, news, and the web). Consequently, the generalizability of our findings to other languages and more diverse or specialized domains remains to be thoroughly tested. 

Furthermore, the model's internal mechanisms may be susceptible to certain weaknesses. One potential issue is an over-reliance on specific Part-of-Speech (POS) patterns for keyword identification. The model might learn heuristic shortcuts, associating keywords primarily with common grammatical structures (e.g., noun-noun compounds or adjective-noun phrases). This could limit its ability to identify valid but unconventional keywords and may lead to false positives. A related concern, specific to the MoE architecture, is the risk of the expert routing mechanism overfitting to superficial cues. The router might learn to dispatch tokens based on shallow features like capitalization or token position rather than deep semantic content, which would prevent true expert specialization and harm the model's robustness on out-of-distribution data.

From a computational and environmental standpoint, the introduction of the BiLSTM encoder in the custom token classification head increases the model's complexity. This makes both inference and fine-tuning slightly slower and more computationally demanding than the vanilla DeBERTa model. While the performance difference was negligible in our experiments, it could become a significant factor when scaling to much larger datasets. More broadly, the training of large models like DeBERTa is an energy-intensive process with an associated environmental cost, a factor that is critical to consider in the development of ever-larger architectures.

Methodologically, our specialization analysis relied on spaCy for Named Entity Recognition. As a general-domain NLP tool, its performance can be suboptimal on specialized texts like scientific articles. This reliance might have affected the reliability of the analysis by yielding fewer or less accurate named entities, potentially skewing the correlation results.

Finally, it is crucial to consider the broader implications and potential for misuse. A model effective at automated keyword extraction could be exploited for malicious purposes, such as "black-hat" Search Engine Optimization (SEO). Such a system could be used to generate spammy keywords to artificially boost the ranking of low-quality content, thereby degrading the integrity of information retrieval systems.

\section{Ethics Statement}

This work does not pose any ethical issues. All the data and tools used in this paper are publicly available under the MIT license. No private data or non-public information is used in this work.

\section*{Acknowledgments}

The authors acknowledge the financial support from the Slovenian Research Agency for research core funding (No. P2-0103) and projects Embeddings-based techniques for Media Monitoring Applications (EMMA, No. L2-50070) and Large Language Models for Digital Humanistics (LLM4DH, No. GC-0002). A Young Researcher Grant PR-12394 supported the work of the last author.

\bibliography{latex/bibilography}

\appendix

\section{Hyperparameters Used}
\label{sec:appendix-hyperparameters}

We use the same hyperparameter setting across all datasets, namely, the following values were used during fine-tuning: learning rate of 2e-4, LoRA intrinsic rank of 16, LoRA alpha of 16, LoRA dropout of 0.1 sequence length of 256, 4 experts, and the top k value of 2.

\section{Model Size and Budget}
\label{sec:appendix-budget}

The final model with MoE and BiLSTM layers, which uses the ``deberta-v3-base'' backbone (available at \url{https://huggingface.co/microsoft/deberta-v3-base} under the MIT license) has 274M parameters, out of them 45M are trainable during LoRA fine-tuning. 

We used two NVIDIA A100 80GB GPUs in our experiments. The total computational budget for all experiments, i.e., validation set training and full training with five random seeds and four different experimental settings (feed-forward classification head, just MoE layer, just BiLSTM layer, both MoE and BiLSTM layers), was 1400 GPU hours.

\section{Examples of Keyword Identification}
\label{sec:appendix-examples}

This section presents examples of extracted keywords on randomly selected instances from all datasets.

\doc{Inspec}{The Bagsik Oscillator without complex numbers. We argue that the analysis of the so-called Bagsik Oscillator, recently published by Piotrowski and Sladkowski (2001), is erroneous due to: (1) the incorrect banking data used and (2) the application of statistical mechanism apparatus to processes that are totally deterministic.}

\pred{incorrect banking data, statistical mechanism apparatus, bagsik oscillator}
\true{data, statistical mechanism apparatus, complex numbers, bagsik oscillator}

\doc{Krapivin}{Solving Equations in the Relational Algebra. Enumerating all solutions of a relational algebra equation is a natural and powerful operation which, when added as a query language primitive to the nested relational algebra, yields a query language for nested relational databases, equivalent to the well-known powerset algebra.  We study sparse equations, which are equations with at most polynomially many solutions.  We look at their complexity and compare their expressive power with that of similar notions in the powerset algebra.}

\true{relational algebra, nested relation, equation}
\pred{equations, relational algebra, powerset}

\doc{KP20k}{An algebraic approach to guarantee harmonic balance method using grobner base. Harmonic balance (HB) method is well known principle for analyzing periodic oscillations on nonlinear networks and systems. Because the HB method has a truncation error, approximated solutions have been guaranteed by error bounds. However, its numerical computation is very time-consuming compared with solving the HB equation. This paper proposes an algebraic representation of the error bound using Grobner base. The algebraic representation enables to decrease the computational cost of the error bound considerably. Moreover, using singular points of the algebraic representation, we can obtain accurate break points of the error bound by collisions.}

\true{algebraic representation, singular point, error bound, grobner base, harmonic balance method}
\pred{grobner base, bound, harmonic balance, harmonic balance method}

\doc{KPTimes}{Afghan Police Chief Is Killed as He Tries to Turn Tide Against Taliban. A hard-charging Afghan police chief with deep experience in Afghanistan's long conflict with the Taliban was killed in a blast on Sunday in the country's eastern Nangarhar Province, which has been under threat from the Taliban and affiliates of the Islamic State. The police chief, Gen. Zarawar Zahid, was visiting an outpost in the Hisarak district when explosives placed near the outpost detonated, according to Attaullah Khogyani, a spokesman for the governor of Nangarhar. One of General Zahid's bodyguards was wounded, Mr. Khogyani said. More Than 14 Years After U.S. Invasion, the Taliban Control Large Parts of Afghanistan At least one-fifth of the country is controlled or contested by the Taliban. The Taliban claimed responsibility for the killing, according to a statement by the insurgency's spokesman, Zabiullah Mujahid. The attack came a week after twin bombings outside Afghanistan's Ministry of Defense killed at least 40 people , including several senior security officials. Nangarhar, which borders Pakistan, has faced mounting security perils over the past couple of years, with new Islamic State affiliates complicating the threat from the Taliban. Zabihullah Zmarai, a member of the provincial council, said the Islamic State posed a danger in five districts, despite repeated operations by the Afghan Army. Out of the 22 districts, only six are secure, he said. The Taliban's presence across nearly a dozen districts varies, Mr. Zmarai said. But the Hisarak district faced a collapse in recent weeks. That drew the attention of General Zahid, who had gone there to supervise a counterattack. Over the past decade, he rose from a bodyguard to a well-regarded police chief of several volatile provinces. His postings included two stints as the police chief of southeastern Ghazni Province, and one term each in Zabul and Paktika Provinces. General Zahid was seen as a hands-on commander, often arriving at the front lines unannounced. When a major cultural event drew world leaders to the ancient city of Ghazni, the general was photographed riding around the city on the back of a motorcycle to check on security measures. He had been wounded twice and had lost two brothers during the decades of war in Afghanistan. In June, he took part in clashes with Pakistani forces that erupted on the border. In a Facebook video that he posted, he appeared beside two mortars and shouted to his men, "Strike hard enough to blow up Nawaz Sharif's home," referring to the prime minister of Pakistan. Sediq Sediqqi, the spokesman for Afghanistan's Ministry of Interior Affairs, called General Zahid "one of the bravest commanders of Afghan police.cHe lost his life on the front line of duty in the fight against terrorism," Mr. Sediqqi said.}

\true{afghanistan, taliban, zarawar zahid, bombs, nangarhar province,terrorism}
\pred{afghanistan, zarawar zahid, taliban}

\doc{JPTimes}{Photo report: FOODEX Japan 2013. FoodEx is the largest trade exhibition for food and drinks in Asia, with about 70,000 visitors checking out the products presented by hundreds of participating companies. I was lucky to enter as press; otherwise, visitors must be affiliated with the food industry and pay to enter. The FoodEx menu is global, including everything from cherry beer from Germany and premium Mexican tequila to top-class French and Chinese dumplings. The event was a rare chance to try out both well-known and exotic foods and even see professionals making them. In addition to booths offering traditional Japanese favorites such as udon and maguro sashimi, there were plenty of innovative twists, such as dorayaki , a sweet snack made of two pancakes and a red-bean filling, that came in coffee and tomato flavors. While I was there I was lucky to catch the World Sushi Cup Japan 2013, where top chefs from around the world were competing and presenting a wide range of styles that you would not normally see in Japan, like the flower makizushi above.}

\true{foodex}
\pred{foodex, japan, food}

\doc{openKP}{"Quickly create pointandclick games and virtual tours for Windows native PSP iPhone and iPod Touch web apps No programming required very easy to use Free edition contains all the main features Includes free drawing tool and music composer NEW Create Games for iPhone DOWNLOAD FREE GAMES CREATED WITH ADVENTURE MAKER WATCH THE GUIDED TOUR Whats New in version 45 Whats New in version 44 }

\true{virtual tours, adventure maker, no programming required}
\pred{virtual tours}

\section{Prompt Used for LLM-based Keyword Extraction}
\label{sec:prompts}
For each document processed in our experiments, we applied the same prompt to ensure consistency in the extraction of relevant keywords. We condition the generation to return valid json prompts, so we can uniformly parse and compare the results. The prompt used for both the  GPT4o-mini and Gemini-1.5-flash-8B models is provided below:

\begin{WrappedVerbatim}
You are an assistant that specializes in text analysis. Your task is to extract the most relevant keywords from the following document. 
Extract only keywords present in the document. Please analyze the text and provide a comma-separated list of keywords ordered from most important to least important.
Return a maximum of 10 keywords.

Document:
\"\"\"{document}\"\"\"

Return only the list of keywords, formatted as a json list (i.e. with square brackets [ and ]).
\end{WrappedVerbatim}

\section{Performance of our approaches trained solely on training sets}

To make a comparison between our approach and other baselines (i.e., especially non-sequence labelling supervised baselines, which were not fine-tuned on the validation data) more comprehensive, Table \ref{tbl:results-just-pretraining} presents results when training of our approaches is done solely on KP20k, KPTimes, and OpenKP, with no validation data. Note that in this training setting our approaches outperform all baselines in 7 out of 13 cases. On the other hand, a large drop in performance can be observed for all our approaches on the Inspec dataset, where the keyword labelling regime is very different from the labelling regime of the KP20k training set, on which the models were trained.

\begin{table*}[h!]
    \centering
    \resizebox{1.0\textwidth}{!}{
    \begin{tabular}{c|cc|cc|cc|cc|cc|ccc}
    \toprule
    \midrule
    \multirow{3}{*}{}& \multicolumn{2}{c|}{\underline{KP20k}} & \multicolumn{2}{c|}{\underline{Inspec}} & \multicolumn{2}{c|}{\underline{Krapivin}} & \multicolumn{2}{c|}{\underline{KPTimes}} & \multicolumn{2}{c}{\underline{JPTimes}} & \multicolumn{3}{c}{\underline{openKP}} \\ 
    & \bf F1@5 & \bf F1@10 & \bf F1@5 & \bf F1@10 & \bf F1@5 & \bf F1@10 & \bf F1@5 & \bf F1@10 & \bf F1@5 & \bf F1@10 & \bf F1@1 & \bf F1@3 & \bf F1@5 \\
    \midrule \midrule
    \multicolumn{14}{c}{\textbf{ Unsupervised baselines }} \\\midrule \midrule
    TF-IDF & 7.2 & 9.4 & 16.0 & 24.4 & 6.7 & 9.3 & 17.9 & 15.1 & 26.6 & 22.9 & 19.6 & 22.3 & 19.6 \\
    TextRank & 18.0 & 15.0 & 28.6 & 33.9 & 18.5 & 16.0 & 2.2 & 3.0 & 1.2 & 2.6 & 5.4 & 7.6 & 7.9 \\
    \midrule \midrule \multicolumn{14}{c}{\textbf{ Large Language Model baselines }} \\\midrule \midrule
    ChatGLM2-6b & / & / & 25.1 & 30.1 & / & / & / & / & / & / & 16.0 & 11.0 & 8.6 \\

    ChatGPT & 23.2 & 10.8 & 35.2 & 33.9 & 12.8 & 11.7 & 27.9 & / & / & / & 20.8 & 20.4 & 16.6 \\ 
    GPT4o-mini & 22.7 & 19.8 & 34.4 & 42.1 & 22.1 & 20.5 & 19.5 & 16.6 & 29.3 & 25.2  & 34.0 & 26.8 & 26.8 \\ 
    Gemini-1.5-flash-8B & 26.5 & 23.6 & 35.9 & 42.1 & 26.6 & 25.9 & 20.4 & 26.9  & 31.8 & 27.5 & 33.1 & 25.5 & 25.6 \\
    \midrule \midrule \multicolumn{14}{c}{\textbf{ Supervised baselines }} \\\midrule \midrule
    CopyRNN & 31.7 & 27.3 & 24.4 & 28.9 & 30.5 & 26.6 & 40.6 & 39.3 & 25.6 & 24.6 & 21.7 & 23.7 & 21.0 \\
    CatSeqD & 34.8 & 29.8 & 27.6 & 33.3 & 32.5 & 28.5 & 42.4 & 42.4 & 23.8 & 23.8 & / & / & / \\
    GPT-2 & 27.5 & 27.8 & 41.3 & 46.9 & 25.3 & 25.3 & 42.1 & 42.3 & 33.1 & 33.6 & / & / & / \\
    GPT-2+BiLSTM-CRF & 35.5 & 36.0 & 46.2 & 52.4 & 28.7 & 28.8 & 47.8 & 47.9 & 38.6 & 38.9 & / & / & / \\
    TNT-KID & 33.6 & 33.8 & 46.0 & 53.6 & 31.0 & 32.0 & 48.5 & 48.5 & 35.9 & 36.1 & / & / & / \\
    JointKPE & \textbf{41.7} & 34.4 & 35.2 & 35.0 & \textbf{36.0} & 29.2 & / & / & / & / & 36.4 & 39.1 & 33.8 \\
    Diff-KPE & \textbf{41.7} & 34.3 & 32.3 & 35.0 & 35.0 & 31.4 & / & / & / & / & \textbf{37.8} & 38.5 & 32.7 \\
    HybridMatch & / & / & / & / & / & / & / & / & / & / & 37.3 & 39.4 & 34.1 \\
    \midrule \midrule
    \multicolumn{14}{c}{\textbf{ Our approaches - just pre-training }} \\ \midrule \midrule
    DeBERTa & 35.2 & 35.3 & 14.5 & 14.5  & 28.8 & 28.9 & 55.6 & 55.7 & 37.4 & 37.5 & 34.1 & \textbf{42.9} & \textbf{43.2} \\
    DeBERTa MoE & 33.9 & 34.0 & 14.7 & 14.8 & 28.5 & 28.8 & 56.4 & 56.5 & 39.1 & 39.3 & 34.7 & 42.7 & 43.0 \\
    DeBERTa RNN & 33.6 & 33.7 & 13.9 & 13.9 & 27.5 & 27.7 & 53.2 & 53.2& 36.8 & 36.9 & 33.4 & 39.7 & 39.8 \\
    DeBERTa MoE RNN & 36.3 & \textbf{36.5} & 15.1 & 15.1 & 29.6 & 30.3 & \textbf{59.7} & \textbf{59.8} & \textbf{39.3} & \textbf{39.5} & 29.6 & 41.5 & 42.2\\
    \midrule
    \bottomrule
    \end{tabular}
    }
     \caption{Performance comparison between our approaches trained solely on KP20k, KPTimes, and OpenKP, and the baseline approaches. Only one seed is used in this scenario. Bold represents the best score according to a specific criterion on a specific dataset.}
    \label{tbl:results-just-pretraining}
\end{table*}

\end{document}